\documentclass[conference]{IEEEtran}
\IEEEoverridecommandlockouts
\usepackage{cite}
\usepackage{amsmath,amssymb,amsfonts}
\usepackage{algorithmic}
\usepackage{graphicx}
\usepackage{textcomp}
\usepackage{xcolor}
\usepackage{bm}
\def\BibTeX{{\rm B\kern-.05em{\sc i\kern-.025em b}\kern-.08em
    T\kern-.1667em\lower.7ex\hbox{E}\kern-.125emX}}
\begin{document}

\title{Triplet Loss for Knowledge Distillation}

\author{\IEEEauthorblockN{Hideki Oki, Motoshi Abe, Jyunichi Miyao and Takio Kurita}
\IEEEauthorblockA{\textit{Department of Information Engineering} \\
\textit{Hiroshima University}\\
Higashi-Hiroshima, Japan \\
\{hidebon9105, i13abemotoshi\}@gmail.com, \{miyao, tkurita\}@hiroshima-u.ac.jp}

}

\maketitle

\begin{abstract}
In recent years, deep learning has spread rapidly, and deeper, larger models have been proposed.
However, the calculation cost becomes enormous as the size of the models becomes larger.
Various techniques for compressing the size of the models have been proposed to improve performance while reducing computational costs.
One of the methods to compress the size of the models is knowledge distillation (KD).
Knowledge distillation is a technique for transferring knowledge of deep or ensemble models with many parameters (teacher model) to smaller shallow models (student model).
Since the purpose of knowledge distillation is to increase the similarity between the teacher model and the student model, we propose to introduce the concept of metric learning into knowledge distillation to make the student model closer to the teacher model using pairs or triplets of the training samples.
In metric learning, the researchers are developing the methods to build a model that can increase the similarity of outputs for similar samples.
Metric learning aims at reducing the distance between similar and increasing the distance between dissimilar.
The functionality of the metric learning to reduce the differences between similar outputs can be used for the knowledge distillation to reduce the differences between the outputs of the teacher model and the student model.
Since the outputs of the teacher model for different objects are usually different, the student model needs to distinguish them.
We think that metric learning can clarify the difference between the different outputs, and the performance of the student model could be improved.
We have performed experiments to compare the proposed method with state-of-the-art knowledge distillation methods.
The results show that the student model obtained by the proposed method gives higher performance than the conventional knowledge distillation methods.
\end{abstract}

\begin{IEEEkeywords}
Convolutional Neural Network, Metric Learning, triplet loss, knowledge distillation
\end{IEEEkeywords}

\section{Introduction}
After the deep Convolutional Neural Network (CNN) proposed by Krizhevsky et al. \cite{Krizhevsky2012} won the ILSVRC 2012 with higher score than the conventional methods, it became very popular for image classification and object recognition. 
Later, with the increase in computer performance, deeper and larger models such as VGG \cite{Simonyan2014} and ResNet \cite{He2016} were proposed.
It is believed that deeper models with more parameters perform better performance than the shallow models.
However, the calculation cost becomes enormous as the size of the models becomes larger.
Various techniques for compressing the size of the models have been proposed to improve performance while reducing computational costs.

One of the methods to compress the size of the models is knowledge distillation (KD).
Knowledge distillation is a technique for transferring knowledge of deep or ensemble models with many parameters (teacher model) to smaller shallow models (student model).
Specifically, the learning of the student model is accelerated by using the output of the trained teacher model.
For example, Ba et al. \cite{Ba2014} used the square error between the teacher model and the student model as the student model loss.
Hinton et al. \cite{Hinton2015} used  the  KL-divergence  of  the  softmax  output  of  both  models  as  the  loss of the student model.

In knowledge distillation, the knowledge of the teacher model is transferred to the student model by making the student model learn to imitate the output of the teacher model.
In other words, knowledge distillation can also be considered as a method for increasing the similarity between the outputs of the teacher model and the student model.
So, in this paper, we propose to introduce metric learning into knowledge distillation.

In metric learning, the researchers are developing the methods to build a model that can increase the similarity of outputs for similar samples.
Representative methods for deep metric learning include the Siamese Network \cite{Bromely1994,Chopra2005,Hadsell2006} and the Triplet Network \cite{Wang2014}, \cite{Hoffer2015}.
These are learned so that the Euclidean distance between outputs for similar samples is reduced.
In addition, learning is performed so that the Euclidean distance between outputs for dissimilar samples becomes large.
That is, there is a function to clarify the difference between dissimilar outputs.
The model will try to get the same output for the same object, and will be able to clearly distinguish the difference between different objects.
With this feature, the metric learning model has performed extremely well in areas such as face verification and human re-identification.

In this paper, we propose to introduce this metric learning concept into knowledge distillation.
The functionality of the metric learning to reduce the differences between similar outputs can be used for the knowledge distillation to reduce the differences between the outputs of the teacher model and the student model.
Since the outputs of the teacher model for different objects are usually different, the student model needs to distinguish them.
We think that metric learning can clarify the difference between the different outputs, and the performance of the student model could be improved.

The effectiveness of the proposed approach is experimentally confirmed for image classification tasks.
Experimental results show that our method dramatically improves the performance of the student model.
Also, we have performed experiments to compare the proposed method with state-of-the-art knowledge distillation methods.
The results show that the student model obtained by the proposed method gives higher performance than the conventional knowledge distillation methods.

\section{Related Works}

\subsection{Deep Convolutional Neural Network}

The deep CNN has been proven to be effective and has been applied many applications such as for image classification\cite{Krizhevsky2012,He2016}, object detection\cite{He2017}, image segmentation\cite{Ronneberger2015} and so on.

The computation within the convolution layers is regarded as a filtering process of the input image as
\begin{align} \label{eq:conv}
f_{p,q}^{(c)}=h(\sum^{convy-1}_{r=0}\sum^{convx-1}_{s=0}w^{(c)}_{r,s}f^{(c-1)}_{p+r, q+s}+b^{(c)}) \; ,
\end{align}
where $w^{(c)}_{r,s}$ is the weight of the neuron indexed as $(r,s)$ in the $c$-th convolution layer and $b^{(c)}$ is the bias of the $c$-th convolution layer. 
The size of the convolution filter is given as $convx \times convy$. 
The activation function of each neuron is denoted as $h$. 
Usually, pooling layers are added after the convolution layers. 
The pooling layer performs downsampling for reducing computational costs and enhancing against micro position changes. 
Fully-connected layers like multi layer perceptron is connected to the convolution layers which is used to construct the classifier.

\subsection{Metric Learning}

Metric learning is a method of learning embedding for metric (Euclidean distance, cosine similarity, etc.).
Metric learning is applied to a wide range of tasks such as image search, biometric authentication, and abnormality detection.

\begin{figure}[ht]
\begin{center}
\includegraphics[width=80mm]{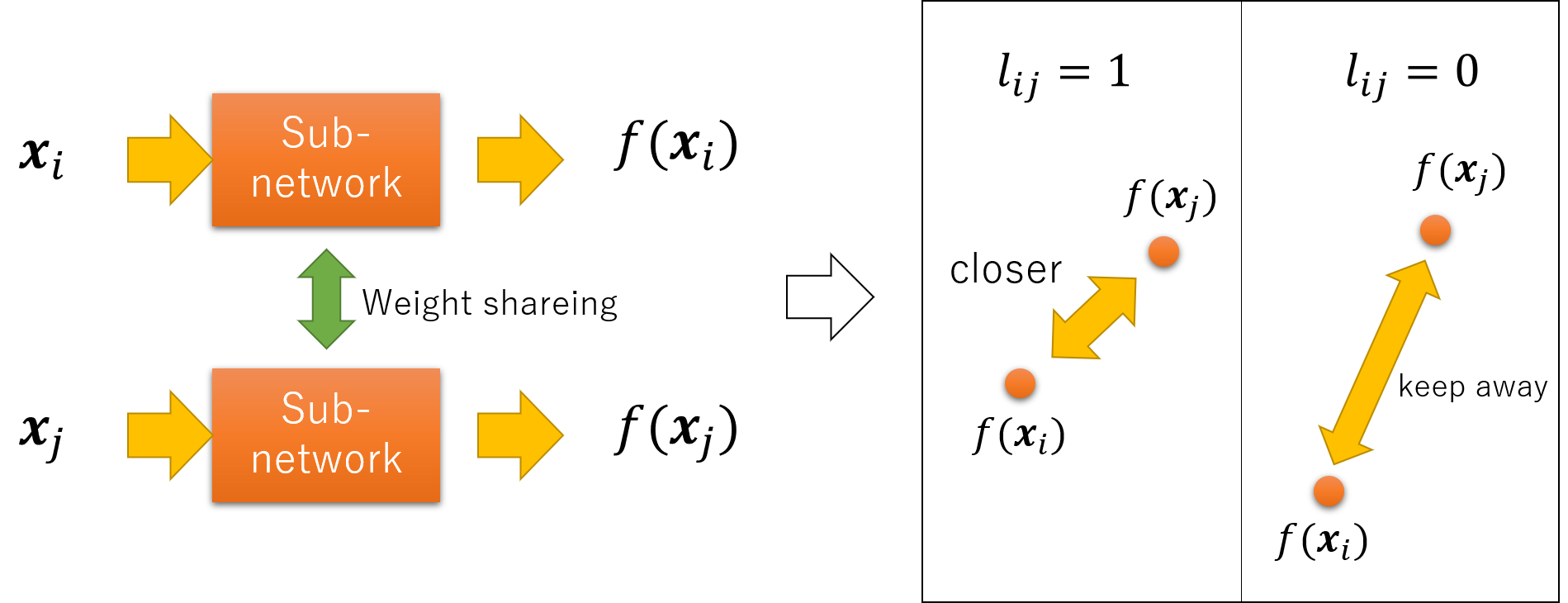}
\caption{The structure of Siamese Network. The distance metric is learned depending on the label $l_{ij}$ given to each sample pair $\bm{x}_i$ and $\bm{x}_j$.}
\label{fig:siamese}
\end{center}
\end{figure}

A typical model for deep learning embedding is the Siamese Neural Network \cite{Bromely1994,Chopra2005,Hadsell2006}.
The Siamese Network consists of two identical networks joined at their outputs.
The two networks extract feature vectors from two different samples.
Usually, the weights of the two networks are shared.
The objective function of the optimization for training the parameters of the networks is defined by using these extracted feature vectors as
\begin{align} \label{eq:contrastive}
E&=\frac{1}{2|\chi^2|}\sum_{(i,j) \in \chi^2} l_{ij}(D_{ij})^2 \notag \\
&  \ \ \ + (1-l_{ij})max(m-D_{ij}, ~0)^2 \; ,
\end{align}
where $D_{ij}$ represents the distance between the pair of the outputs $f(\bm{x}_i)$ and $f(\bm{x}_j)$ of each network for the sample pair $\bm{x}_i$ and $\bm{x}_j$.
The distance $D_{ij}$ is defined by
\begin{align} 
D_{ij}&=||f({\bm x}_{i})-f({\bm x}_{j})||_2 \; ,
\label{eq:dis}
\end{align}
where $m$ is a parameter indicating the distance between clusters and 
$\chi^2$ is an index set of sample pairs randomly generated from the samples in the mini-batch.
A label $l$ is assigned for each sample pair such that 
label is $l_{ij}=1$ when the pair $i$ and $j$ is similar and label is $l_{ij}=0$ when the pair $i$ and $j$ is dissimilar.
The scheme of Siamese Network is shown in Fig. \ref{fig:siamese}.

After the training of Siamese Network, the distance between the outputs for dissimilar pair will be far while the distance between the outputs for similar pair become close.
The Siamese Network can only consider the metric between samples pairwise.
For this reason, the Siamese Network have to uniquely determine the concept of similarity.
For example, if there are two different male images, in the case of the concept of “gender”, they should be judged to be similar.
However, in the case of the concept of “individuals”, they should be judged not to be similar.
It is difficult to express these multiple concepts in the Siamese Network.

\begin{figure}[ht]
\begin{center}
\includegraphics[width=80mm]{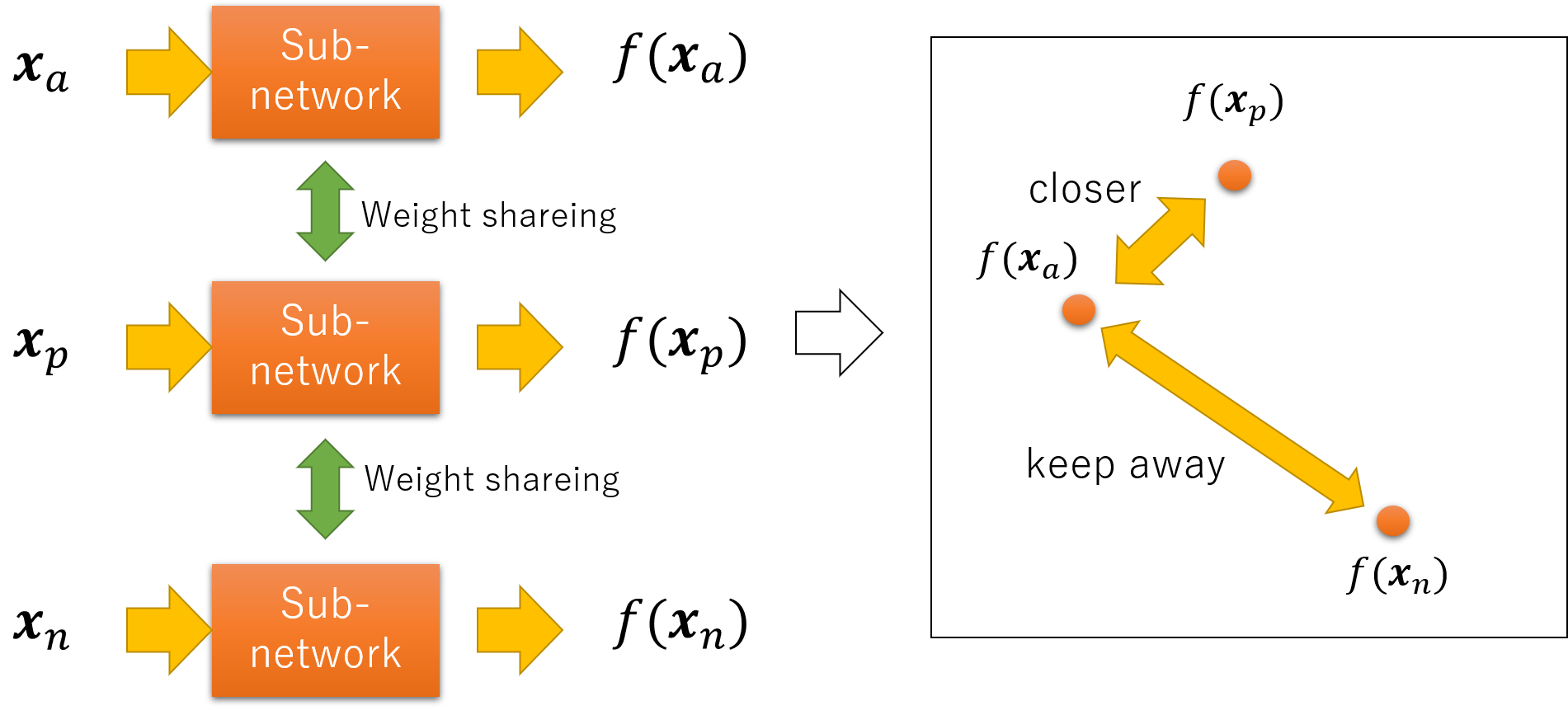}
\caption{The structure of Triplet Network. The distance metric is learned by using three networks with the shared weights from triplet $\bm{x}_a$, $\bm{x}_p$ and $\bm{x}_n$ which are called "anchor", "positine", and "negative". Learning progresses as closer "anchor-positive", and keep away "anchor-negative".}
\label{fig:triplet}
\end{center}
\end{figure}

Wang et al. \cite{Wang2014}, Hoffer et al. \cite{Hoffer2015} proposed the Triplet Network (Figure \ref{fig:triplet}), an extension of the Siamese Network to triplet-wise.

Triplet Network is designed to learn embedding from a triplet of samples called "anchor", "positive" and "negative".
The triplet network learns embedding such that the distance between "anchor" ${\bm x_a}$ and "positve" ${\bm x_p}$ is smaller than the distance between "anchor" ${\bm x_a}$ and "negative" ${\bm x_n}$.
Various losses have been proposed for the Triplet Network.
The triplet loss 
\begin{align} \label{eq:triplet}
    E&=\sum_{(a,p,n) \in \Theta}max(0, m + ||f({\bm x_a})-f({\bm x_p})||_2^2 \notag \\ 
    &- ||f({\bm x_a}) - f({\bm x_n})||_2^2)
\end{align}
is generally used, 
where $f(\cdot)$ denotes the output of the model, and $m$ is a parameter indicating the distance between clusters.
$\Theta$ is an index set of "anchor", "positive", and "negative".
The triplet network learns so that the anchor-positive distance is relatively closer than the anchor-negative distance.
Therefore, multiple similar concepts can be considered without depending on one similar concept.
Triplet Network can overcome the drawbacks of Siamese Network.

\subsection{Knowledge Distillation}

\begin{figure}[ht]
\begin{center}
\includegraphics[width=80mm]{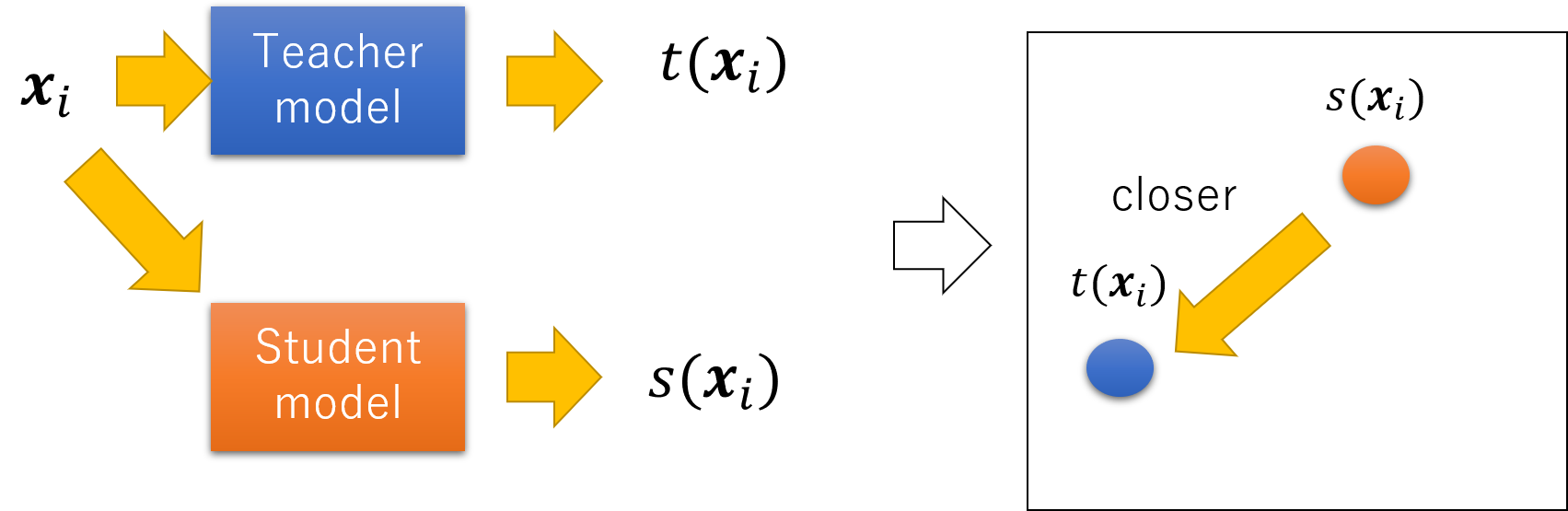}
\caption{The structure of conventional knowledge distillation. The student model tries to mimic the teacher model.}
\label{fig:KD}
\end{center}
\end{figure}
Knowledge distillation is a technique for transferring knowledge of deep or ensemble model with many parameters (teacher model) to smaller shallow model (student model).
Various approaches have been proposed for transferring knowledge.

Ba et al. \cite{Ba2014} considered the difference between the input vectors of the softmax activation function (logits) of the teacher model and the student model as loss (Equation(\ref{eq:Ba})), and proposed to train the student model so that the feature vectors of the student model got close to the teacher model (BKD).
The loss to measure the similarity between the teacher model and the student model is defined by the mean squared errors between the feature vector $t(\cdot)$ of the teacher model and the feature vector $s(\cdot)$ of the student model as 
\begin{align} \label{eq:Ba}
E_{BKD}= \frac{1}{2}\sum_{i \in \chi} ||t(\bm{x}_i) - s(\bm{x}_i)||^2_2,
\end{align}
where $\chi$ is the set of training samples.
As a result, the authors found that the performance of the student model was higher than the network trained by the student model alone.
They claimed that the uncertainty of the feature vectors of the teacher model is useful for learning the student model.

Hinton et al. \cite{Hinton2015} proposed training the student model so that the softmax outputs of the teacher model and the softmax outputs (probability) of the student model are close (HKD).
They used the KL-divergence of the softmax outputs of both models as the loss for training the student model.
The loss based on the KL-divergence is defined as
\begin{align} \label{eq:Hinton}
    E_{HKD} = \sum_{i \in \chi} KL(\mbox{softmax}(\frac{t(\bm{x}_i)}{T}), \mbox{softmax}(\frac{s(\bm{x}_i)}{T})),
\end{align}
where $\mbox{softmax}(\cdot)$ is the softmax function,
$KL(\cdot)$ is KL-divergence between the softmax outputs of the teacher model $\{p_i\}$ and the softmax outputs $\{q_i\}$ of the student model defined by
\begin{align} \label{eq:KL-divergence}
KL(\bm{p}, \bm{q}) = \sum_{i} p_i log(\frac{p_i}{q_i}) \; ,
\end{align}
and $T$ is a temperature parameter.
They argued that negative values in logits could have a positive effect in some cases or negative effect in some cases, on learning.

In recent years, various approaches have been proposed for knowledge distillation.
Park et al. \cite{Park2019} expressed the relationship between the outputs of the teacher model as the Euclidean distance between the two outputs, and transferred it to the student model (RKD-D).
They expressed the similarity between the pair of the outputs $t(\bm{x}_i)$ and $t(\bm{x}_j)$ as
\begin{align} \label{eq:psiD}
    \psi_D(t(\bm{x}_i), t(\bm{x}_j)) = \frac{||t(\bm{x}_i) - t(\bm{x}_j)||_2}{\sum_{(\bm{x}_i, \bm{x}_j) \in \chi^2} ||t(\bm{x}_i) - t(\bm{x}_j)||_2},
\end{align}
where $\chi^2$ is a set of sample pairs randomly generated from the samples in the mini-batch.
They optimized the student model by Huber loss so that the similarity between the outputs of the student model and the similarity of the outputs of the teacher model got closer.
The loss function of RKD-D is defined by using the similarity $\psi_D$ as
\begin{align} \label{eq:RKD-Dloss}
    E_{RKD-D} = \sum_{(\bm{x}_i, \bm{x}_j) \in \chi^2}l(\psi_D(t(\bm{x}_i), t(\bm{x}_j)), \psi_D(s(\bm{x}_i), s(\bm{x}_j))),
\end{align}
where
\begin{eqnarray} \label{eq:Huber}
l(p,q)=\left\{
\begin{array}{ll}
\frac{1}{2}(p-q)^2 & (|p-q| \leq 1) \\
|p-q|-\frac{1}{2} & (otherwise)
\end{array}
\right. \; 
\end{eqnarray} 
is Huber loss.

They also used the cosine of the angle formed by the three outputs as the similarity of the model outputs (RKD-A).
The cosine of the three outputs $t(\bm{x}_i)$, $t(\bm{x}_j)$, and  $t(\bm{x}_k)$ is defined by
\begin{align} \label{eq:psiA}
    &\psi_A^{(t_{ijk})} = \cos \angle t(\bm{x}_i) t(\bm{x}_j) t(\bm{x}_k),
\end{align}
where
\begin{align} \label{eq:psiA-cos}
 \cos \angle t(\bm{x}_i) t(\bm{x}_j) t(\bm{x}_k) &= \langle \bm{e}^{(t_{ij})}, \bm{e}^{(t_{kj})} \rangle
\end{align}
\begin{align}
    \bm{e}^{(t_{ij})} &= \frac{t(\bm{x}_i) - t(\bm{x}_j)}{||t(\bm{x}_i) - t(\bm{x}_j)||_2}, \notag \\ \bm{e}^{(t_{kj})} &= \frac{t(\bm{x}_k) - t(\bm{x}_j)}{||t(\bm{x}_k) - t(\bm{x}_j)||_2} \; .
\end{align}
Loss of RKD-A is also defined by using Huber loss \cite{Huber1992} as
\begin{align} \label{eq:RKD-Aloss}
E_{RKD-A} = \sum_{(\bm{x}_i, \bm{x}_j, \bm{x}_k) \in \chi^3}l(\psi_A^{(t_{ijk})}, \psi_A^{(s_{ijk})}) \: ,
\end{align}
where $\chi^3$ is a set of triplet samples randomly generated from the samples in the mini-batch.

They also argued that using both angles and Euclidean distance for knowledge transfer would further improve the performance of the student model (RKD-DA).
The loss for this case is defined as
\begin{align} \label{eq:RKD-DAloss}
E_{RKD-DA} = \lambda_{RKD-D} E_{RKD-D} + \lambda_{RKD-A} E_{RKD-A} \; ,
\end{align}
where $\lambda_{RKD-D}$ and $\lambda_{RKD-A}$ are the hyper-parameters.

\begin{figure}[ht]
\begin{center}
\includegraphics[width=80mm]{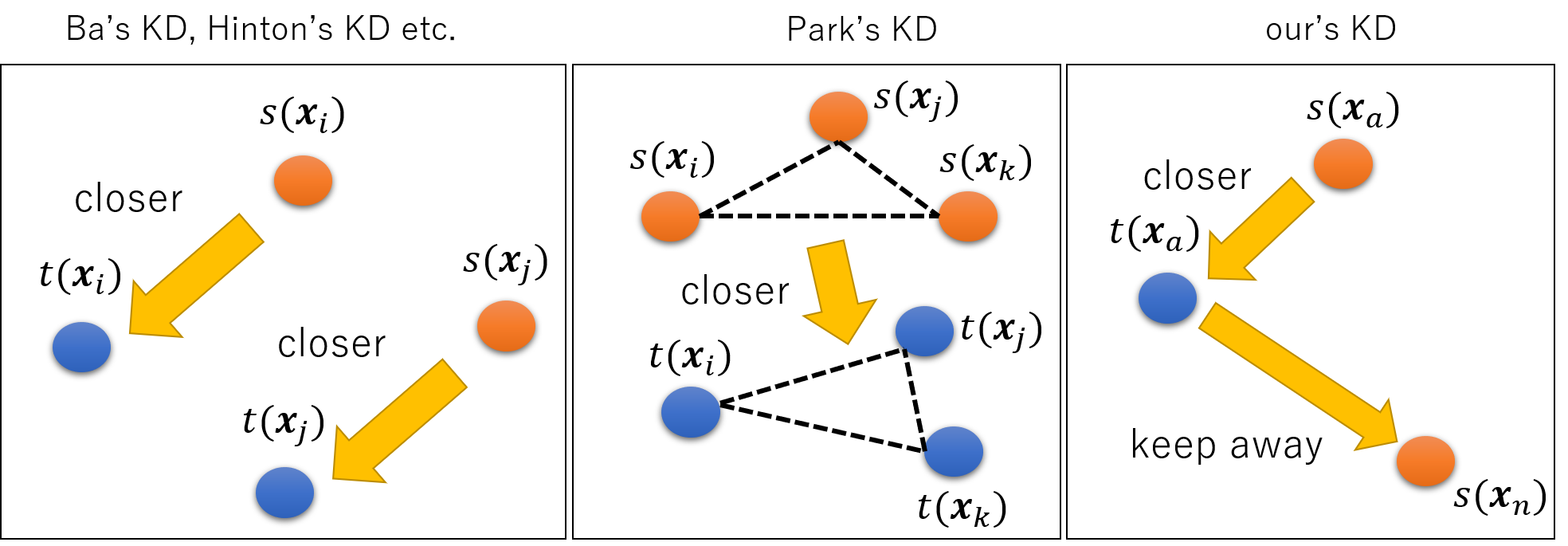}
\caption{The summary of knowledge distillation methods in terms of the loss function.}
\label{fig:loss}
\end{center}
\end{figure}

The pair-wise or triplet-wise similarity between the output of the teacher model to the output of the student model is considered in RKD-D, RKD-A, or RKD-DA while the point-wise transfer was performed in the knowledge distillation methods such as Ba's KD \cite{Ba2014} or Hinton's KD \cite{Hinton2015}.
Thus the knowledge distillation methods such as RKD-D, RKD-A, or RKD-DA succeeded in achieving top performance.

The loss using the output distribution of the teacher model (soft target) can be used alone or combined with the loss using the label of the original training data (hard target).
The combined loss can be defined as
\begin{align} \label{eq:distillation loss}
E_{KD} = E_{hard} + \lambda_{soft} E_{soft},
\end{align}
where $E_{hard}$ is hard target loss and $E_{soft}$ is soft target loss.
In this case, a hyper-parameter $\lambda$ is introduced to keep the balance between the hard target and the soft target.
It is also possible to combine multiple soft target losses.
For example, in Park's KD, they are $\lambda_{RKD-D}$ and $\lambda_{RKD-A}$.

\section{Proposed method}

\begin{figure}[ht]
\begin{center}
\includegraphics[width=80mm]{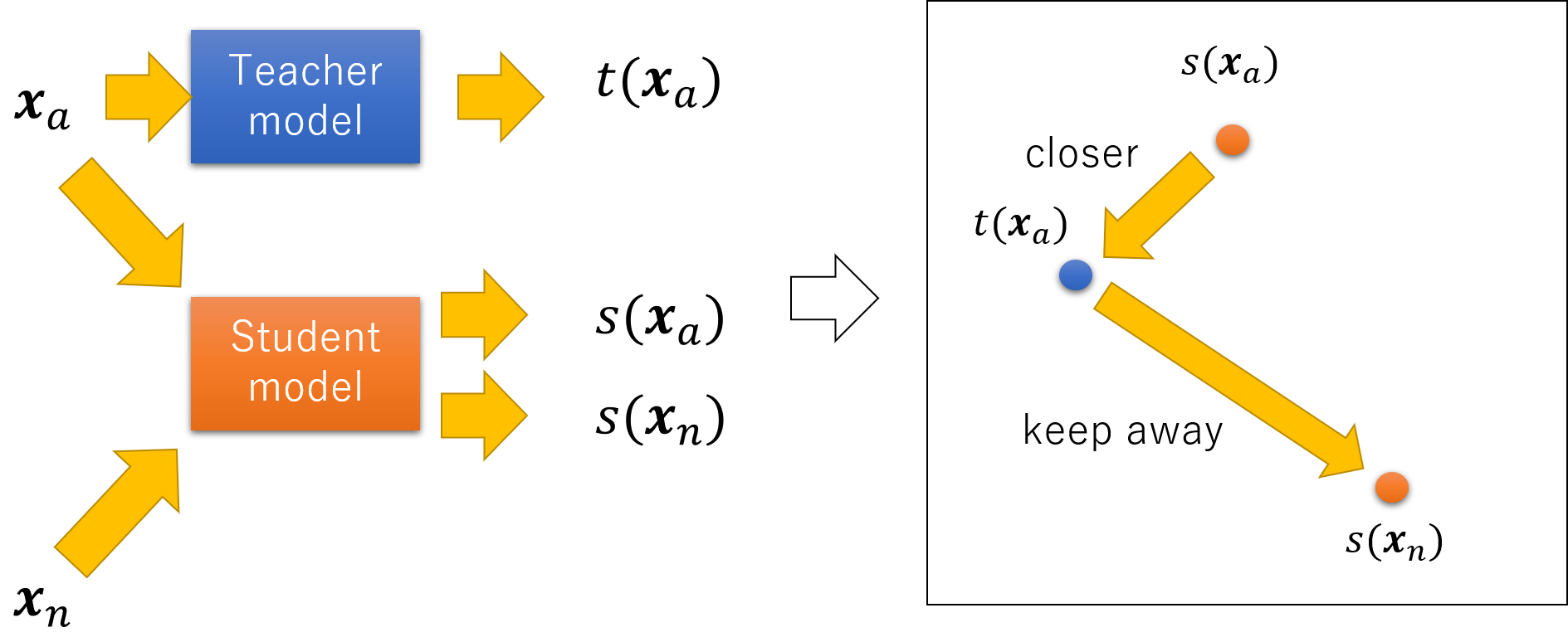}
\caption{The structure of our method. The student model is optimized so that its own output for sample $\bm{x}_a$ and the output of the teacher model for sample $\bm{x}_a$ are closer. At the same time, the student model is optimized so that its own output for sample $\bm{x}_n$ which is the different classes in the soft target and the output of the teacher model for sample $\bm{x}_a$ are keeping away.}
\label{fig:ours}
\end{center}
\end{figure}

Previously proposed methods of knowledge distillation focused on minimizing the difference between the teacher model and the student model in order to transfer knowledge of the teacher model.
For example,  Ba et al. \cite{Ba2014} considered the difference in model output as loss and Hinton et al. \cite{Hinton2015} considered the difference in model output distribution as loss.
In other words, they are solving the optimization problem in which the similarity between the function of the teacher model and the function of the student model is maximized.
There is a possibility that knowledge of the teacher model can be transferred to the student model by using the metric learning method to embed the neighboring relations of the teacher model in the output space of the student model.

Triplet \cite{Hoffer2015}, one of the deep metric learning methods, is a powerful method for learning similarity and dissimilarity (Equation (\ref{eq:triplet})).
Triplet loss has a function to reduce the output distance of "anchor-positive" and a function to increase the output distance of "anchor-negative".
We propose to apply this technique for knowledge distillation.

We define the triplet loss for knowledge transfer as
\begin{align}
    E_{ourKD} &= \sum_{(a,n) \in \Omega} max(0, m + ||t({\bm x_a}) - s({\bm x_a})||_2^2 \notag \\
    &- ||t({\bm x_a}) - s({\bm x_n})||_2^2) \; ,
\end{align}
where $t(\cdot)$ is the output of the teacher model, $s(\cdot)$ is the output of the student model, and $m$ is the margin.
Further, ${\bm x_a}$ is a training sample extracted at random, and ${\bm x_n}$ is a sample classified into a different class from ${\bm x_a}$ in the case of a soft target.
$\Omega$ is the index set of each corresponding sample.
So, the output of the teacher model and the output of the student model for the same sample are considered as "anchor" and "positive", respectively.
Also we consider the output of the student model for samples of different classes in the case of soft target as "negative".
Since we are training the student model, the weight of the teacher model is not updated.
That is, $t(\cdot)$ is treated as a constant during training process.

For our loss, there is a term that makes the outputs of the teacher model and the student model closer. 
This is realized by making the square error of the outputs of the teacher model and the student model a loss.
This is similar to Ba et al. \cite{Ba2014} proposed function.
It is easy to define the loss by using KL-divergence similar with Hinton et al. \cite{Hinton2015}.
In addition, there are terms that increase the distance between outputs of different classes.
In other words, in addition to the conventional loss functions, it is possible to add a constraint that the student model makes the outputs of other classes dissimilar.
This will allow the student model to clarify differences in output between classes.
We have shown experimentally that this proposal contributed to the performance improvement of the student model.
We will also seek further performance improvements in combination with other distillation losses.

\section{experiment}

We experimentally investigated the effectiveness of the proposed method with the image classification task.
CIFAR-10 \cite{Krizhevsky2009} and Tiny ImageNet \cite{Le2015} are used as datasets.

\subsection{Experiments using CIFAR-10}

\begin{table}[ht] 
\caption{model structures for CIFAR-10}
\label{table:structure}
\begin{center}
\begin{tabular}{l|l} 
\hline
student model(5 layers) & teacher model(8 layers)   \\ \hline \hline
conv1(channel:32, filter:3) & conv1(channel:32, filter:3) \\
max pooling(2*2) & batch normalization(32) \\
conv2(channel:32, filter:3) & max pooling(2*2) \\
max pooling(2*2) & conv2(channel:32, filter:3) \\
conv3(channel:64, filter:3) & batch normalization(32) \\
max pooling(2*2) & max pooling(2*2) \\
fully-connected(128) & conv3(channel:64, filter:3) \\
output(10) & batch normalization(64) \\
 & conv4(channel:64, filter:3) \\
 & batch normalization(64) \\
 & conv5(channel:128, filter:3) \\
 &  batch normalization(128) \\
 & max pooling(2*2) \\
 & fully-connected(512) \\
 & dropout \\
 & fully-connected(128) \\
 & dropout \\
 & output(10) \\
\hline
\end{tabular}
\end{center}
\end{table}

CIFAR-10 is a dataset that includes color images of 10 kinds of objects such as "automobile" and "dog."
The size of each image in CIFAR-10 is $32 \times 32$.
The numbers of the training samples and the test samples are 50,000 and 10,000.
We conducted comparative experiments with Ba's KD \cite{Ba2014} (BKD), Hinton's KD \cite{Hinton2015} (HKD), and Park's RKD-DA \cite{Park2019}.

In the case of CIFAR-10, the 8-layers CNN model shown in Table \ref{table:structure} was used as a teacher model, and the 5-layers CNN model was used as a student model.
The number of trainable parameters of the student model is 161,130, and the number of trainable parameters of the teacher model is 1,256,106. 
The number of parameters in the student model is about 12.8$\%$ of the number of parameters in the teacher model.

An activation function ReLU defined by
\begin{align} \label{eq:ReLU}
h(x) = max(0,x)
\end{align}
is used for the outputs of each convolution layer and the outputs of each fully-connected layer except for the last output layer.

In the training, the optimization is done by using the stochastic gradient descent (SGD) with momentum.
The learning rate of SGD is initially set at 0.01 and changed by multiplying 0.1 at every 100 epochs. 
The momentum parameter is set to 0.9.
We also introduced weight decay to prevent over-learning.
We set hard target loss as cross entropy with softmax and combined it with soft target loss.
In addition, we introduced a parameter $\lambda$ as shown in Equation (\ref{eq:distillation loss}) to maintain the balance between hard target loss and soft target loss.
We set $\lambda_{BKD}=2$, $\lambda_{HKD}=16$, $\lambda_{RKD-D}=10$, $\lambda_{RKD-A}=20$ and $\lambda_{oursKD}=2$.
The temperature parameter of Hinton's KD \cite{Hinton2015} was set to 4, and the hyper-parameter $m$ of triplet loss was set to 5.

\begin{table}[ht]
\caption{Classification accuracy for CIFAR-10}
\label{table:cifar-10}
\begin{center}
\begin{tabular}{|l|l|}
\hline
method & accuracy \\ \hline \hline
Student model & $74.66\%$ \\ \hline
Ba's KD & $80.40\%$ \\ \hline
Hinton's KD & $79.16\%$ \\ \hline
RKD-DA &       $79.21\%$     \\ \hline
Ours KD & $\bm{81.14}\%$ \\ \hline \hline
Teacher model  & $83.42\%$     \\
\hline
\end{tabular}
\end{center}
\end{table}

The Table \ref{table:cifar-10} shows the classification accuracy of the student model for test samples in each method.
From the table, it can be seen that in the case of CIFAR10, the proposed method achieves higher performance than the other methods.

\subsection{Experiments using Tiny ImageNet}

Also, we have performed experiments using the Tiny ImageNet dataset.
Tiny ImageNet dataset includes color images of 200 kinds of objects.
The image size is $64 \times 64$, and it has 500 train images and 50 test images per class.
For the case of Tiny ImageNet, VGG19 with batch normalization and VGG11 were used as the teacher model and the student model, respectively.
The numbers of trainable parameters of the teacher model and the student model are 46,028,808 and 35,213,896, respectively.
This means that the number of parameters of the student model is about 76.5$\%$ of those of the teacher model.
We conducted comparative experiments with Ba's KD \cite{Ba2014} (BKD), Hinton's KD \cite{Hinton2015} (HKD), and Park's RKD-DA \cite{Park2019}.

The weights in the convolution layers were pre-trained by using ImageNet, and they were used as the initial weights of the training.
Similarly, the SGD with momentum was used for the optimization.
The learning  rate  of  SGD  with momentum is initially set at 0.001 and multiplied by 0.9  every 3 epochs.  
The  momentum parameter  is  set  to  0.9. 
We also used mixup \cite{Zhang2017} to prevent over-learning.
 Cross-entropy with softmax was used as the hard target loss and was combined with the soft target loss.
Also, we introduced a parameter $\lambda$ as shown in Equation (\ref{eq:distillation loss}) to maintain the balance between hard target loss and soft target loss.
We set $\lambda_{BKD}=2$, $\lambda_{HKD}=16$, $\lambda_{RKD-D}=25$, $\lambda_{RKD-A}=50$ and $\lambda_{oursKD}=2$.
The temperature parameter of Hinton's KD \cite{Hinton2015} was set to 4, and the parameter m of triplet loss was set to 5.

\begin{table}[ht]
\caption{Classification accuracy for Tiny ImageNet}
\label{table:imagenet}
\begin{center}
\begin{tabular}{|l|c|}
\hline
method & accuracy \\ \hline \hline
Student model & $58.63\%$ \\ \hline
Ba's KD & $59.80\%$ \\ \hline
Hinton's KD & $59.80\%$ \\ \hline
RKD-DA & $59.73\%$     \\ \hline
Ours KD & $\bm{60.00}\%$ \\ \hline \hline
Teacher model  & $63.17\%$    \\
\hline
\end{tabular}
\end{center}
\end{table}

Table \ref{table:imagenet} shows the classification accuracy of the student model for the Tiny ImageNet dataset in each method.
The results show that the proposed method gives slightly better performance than the other methods.

\subsection{Experiments on the combined loss}

Park et al. \cite{Park2019} succeeded in achieving even higher performance by combining their method (RKD-DA) with Hinton's method \cite{Hinton2015}.
Here we investigate the effectiveness of the combination of different loss functions.
The loss of each method is combined with Hinton's loss as
\begin{align} \label{eq:distillation loss2}
E = E_{hard} + \lambda_{soft} E_{soft} + \lambda_{HKD} E_{HKD}
\end{align}.
Depending on the datasets CIFAR-10 and Tiny ImageNet, the same network architectures with the previous subsections were used.
The student models were trained, and the classification performances of the trained student models are calculated.

\begin{table}[ht]
\caption{Classification accuracy for combined with HKD}
\label{table:combined}
\begin{center}
\begin{tabular}{|l|c|c|}
\hline
method & CIFAR10 & Tiny ImageNet \\ \hline \hline
Student model &$74.66\%$  & $58.63\%$ \\ \hline
Hinton's KD & $79.16\%$ & $59.80\%$  \\ \hline
Ba's KD + HKD & $80.33\%$ & $60.19\%$  \\ \hline
RKD-DA + HKD &  $79.65\%$   & $59.89\%$      \\ \hline
Ours KD + HKD & $\bm{80.93}\%$ & $\bm{60.62}\%$  \\ \hline \hline
Teacher model  & $83.42\%$  & $63.17\%$  \\
\hline
\end{tabular}
\end{center}
\end{table}

Table \ref{table:combined} shows the classification performance of the test samples for each dataset.
From this table \ref{table:combined}, it is noticed that the classification accuracy of Park's KD \cite{Park2019} was improved by combining with the Hinton's loss for CIFAR-10 dataset, but the classification accuracy of the proposed method and Ba's KD \cite{Ba2014} was not improved.
However the proposed method still gives the best classification accuracy.
For the Tiny ImageNet dataset, all methods showed an increase in the classification accuracy when they are combined with Hinton's KD \cite{Hinton2015}.
Also, the proposed method gives the best accuracy for this case.

Also we consider a combination with Park's KD \cite{Park2019}.
We investigate the effects of combining RKD-DA loss with our loss and we investigate combinations with both Hinton's KD \cite{Hinton2015} and RKD-DA.
Depending on the datasets CIFAR-10 and Tiny ImageNet, the same network architectures with the previous subsections were used.

\begin{table}[ht]
\caption{Classification accuracy of ours KD for combined with RKD-DA}
\label{table:combinedRKD}
\begin{center}
\begin{tabular}{|l|c|c|}
\hline
method & CIFAR10 & Tiny ImageNet \\ \hline \hline
Student model &$74.66\%$  & $58.63\%$ \\ \hline
RKD-DA & $79.21\%$  & $59.73\%$   \\ \hline
ours KD & $\bm{81.14}\%$ & $60.00\%$   \\ \hline
ours KD + RKD-DA & $80.17\%$    &  $\bm{60.32}\%$     \\ \hline
ours KD + HKD + RKD-DA &  $80.30\%$   & $\bm{60.65}\%$      \\ \hline
Teacher model  & $83.42\%$  & $63.17\%$  \\
\hline
\end{tabular}
\end{center}
\end{table}

Table \ref{table:combinedRKD} shows the classification accuracy of the student model in each method.
In the case of the CIFAR10 dataset, no performance improvement was obtained even when combined with RKD-DA.
In the case of the Tiny ImageNet dataset, we succeeded in improving performance by combining it with RKD-DA.
In addition, using both RKD-DA and Hinton's KD in combination with our method resulted in further performance improvements.
Although the combination of our method and Hinton's KD had improved the performance, the combination of both Hinton's KD and RKD-DA achieved even higher performance.

\section{conclusion}
In this paper, we applied the concept of metric learning to knowledge distillation.
By introducing the concept of metric learning, we transfer not only the teacher model knowledge to be imitated but also the knowledge that should not be imitated to the student model.
Our experimental results showed that this approach can significantly improve the performance of the student model.
Also, the combination with Hinton's KD \cite{Hinton2015} and Park's KD \cite{Park2019} succeeded in giving the student model even better performance.
This fact indicates that metric learning contributes to transferring the knowledge of the teacher model to the student model.

On the other hand, the concept of metric learning, such as Triplet loss, has a problem with hard sampling.
In hard sampling, we intentionally select samples that are difficult to learn to get consistent training results.
If you use only samples that are easy to learn, the performance of learning decreases.
In the proposed approach, it is necessary to use hard sampling because the proposed method is based on the Triplet loss.
Thus we have to introduce a way to address this problem, like Hermans et al. \cite{Hermans2017}.

\section*{Acknowledgment}

This work was partly supported by JSPS KAKENHI Grant Number 16K00239.

\bibliographystyle{unsrt}
\bibliography{conference_101719.bbl}

\end{document}